\documentclass[runningheads]{llncs}

\usepackage{eccv}

\usepackage{eccvabbrv}
\usepackage{multibib}
\usepackage{graphicx}
\usepackage{booktabs}
\usepackage{color,soul}
\usepackage{colortbl}
\usepackage{todonotes}
\usepackage{tcolorbox}
\usepackage{graphicx}
\usepackage{booktabs}
\usepackage{multirow}
\usepackage{stackengine}
\usepackage{wrapfig}
\usepackage{enumitem}

\definecolor{mygreen}{HTML}{D1FFBD}
\newcolumntype{a}{>{\columncolor{mygreen}}r}

\usepackage[accsupp]{axessibility}  

\usepackage{hyperref}

\usepackage{orcidlink}

\begin{document}

\title{Getting it \textit{Right}: Improving Spatial Consistency in Text-to-Image Models} 



\author{
Agneet Chatterjee \thanks{Equal contribution. Correspondence to \href{mailto:agneet@asu.edu}{agneet@asu.edu}}\inst{1}\orcidlink{0000-0002-0961-9569} \and
Gabriela Ben Melech Stan \textsuperscript{$\star$} \inst{2}\orcidlink{0000-0001-6893-6647} \and
Estelle Aflalo\inst{2}\orcidlink{0009-0009-2860-6198} \and  \\
Sayak Paul\inst{3}\orcidlink{0000-0003-0217-0778} \and
Dhruba Ghosh \inst{4}\orcidlink{0000-0002-8518-2696} \and 
Tejas Gokhale \inst{5}\orcidlink{0000-0002-5593-2804} \and 
Ludwig Schmidt \inst{4} \and \\
Hannaneh Hajishirzi \inst{4}\orcidlink{0000-0002-1055-6657} \and 
Vasudev Lal\inst{2}\orcidlink{0000-0002-5907-9898} \and 
Chitta Baral\inst{1}\orcidlink{0000-0002-7549-723X} \and 
Yezhou Yang \inst{1}\orcidlink{0000-0003-0126-8976}
}

\authorrunning{A. Chatterjee et al.}

\institute{Arizona State University \and Intel Labs \and Hugging Face \and University of Washington \and University of Maryland, Baltimore County}

\maketitle
\begin{abstract}
  One of the key shortcomings in current text-to-image (T2I) models is their inability to consistently generate images which faithfully follow the spatial relationships specified in the text prompt. 
  In this paper, we offer a comprehensive investigation of this limitation, while also developing datasets and methods that support algorithmic solutions to improve spatial reasoning in T2I models.
  We find that spatial relationships are under-represented in the image descriptions found in current vision-language datasets.
  To alleviate this data bottleneck, we create SPRIGHT, the first spatially focused, large-scale dataset, by re-captioning 6 million images from 4 widely used vision datasets and through a 3-fold evaluation and analysis pipeline, show that SPRIGHT improves the proportion of spatial relationships in existing datasets.
  We show the efficacy of SPRIGHT data by showing that using
  only $\sim$0.25\% of SPRIGHT results in a 22\% improvement in generating spatially accurate images while also improving FID and CMMD scores. 
  We also find that training on images containing a larger number of objects leads to substantial improvements in spatial consistency, including state-of-the-art results on T2I-CompBench with a spatial score of 0.2133, by fine-tuning on \textit{<500} images. 
  Through a set of controlled experiments and ablations, we document additional findings that could support future work that seeks to understand factors that affect spatial consistency in text-to-image models. Project page : \url{https://spright-t2i.github.io/}

  \keywords{Text to Image Generation \and Spatial Relationships}
\end{abstract}
\begin{figure}[t]
    \centering 
    \includegraphics[width=\linewidth]{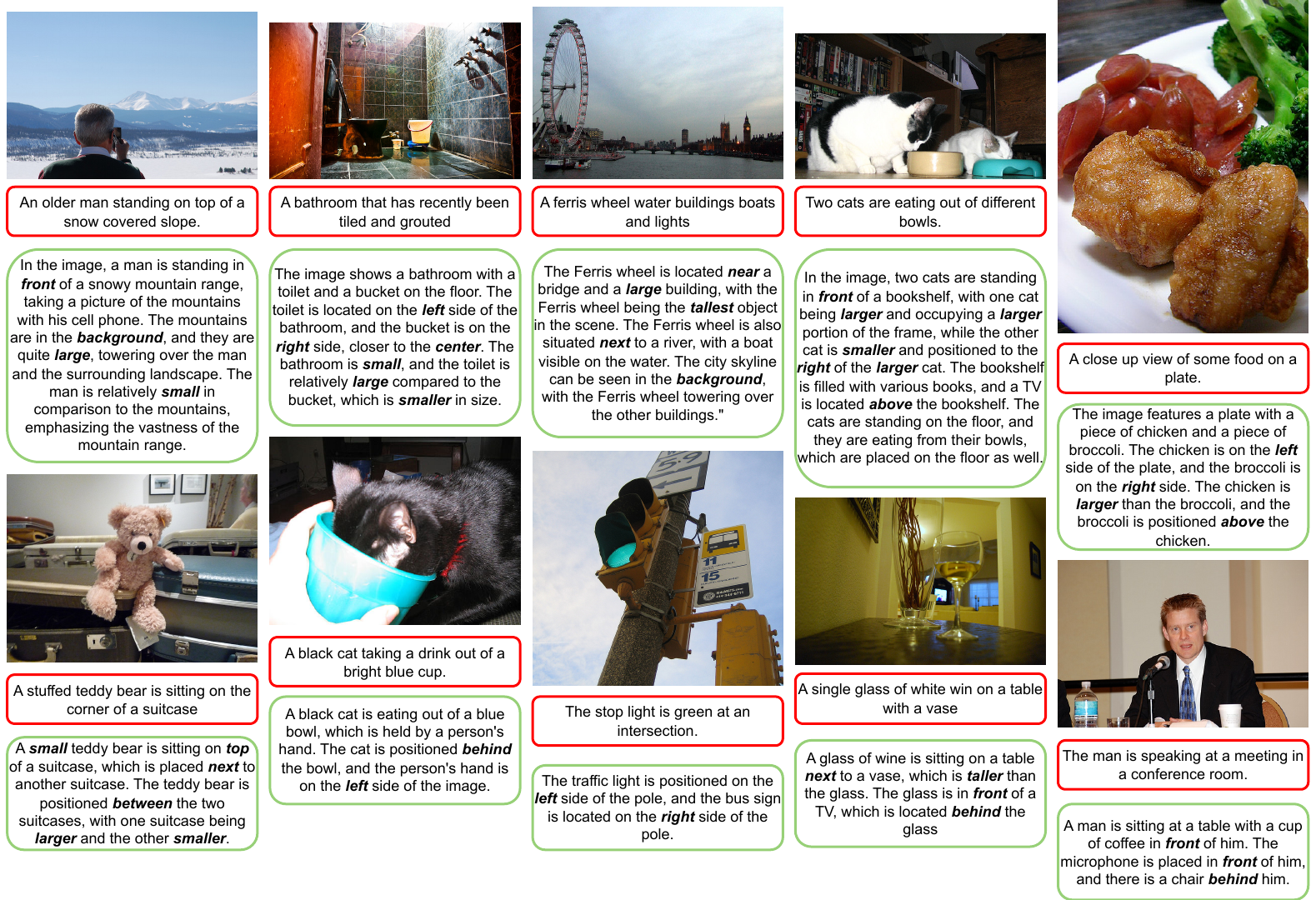}
    \caption{We find that existing vision-language datasets do not capture spatial relationships well. To alleviate this shortcoming, we synthetically re-caption $\sim$6M images with a spatial focus, and create the SPRIGHT (\textbf{SP}atially \textbf{RIGHT}) dataset. Shown above are samples from the COCO Validation Set, where text in \textcolor{red}{red} denotes ground-truth captions and text in \textcolor{ForestGreen}{green} are corresponding captions from SPRIGHT.}
    \label{fig:teaser}
\end{figure}

\section{Introduction}

The development of text-to-image (T2I) diffusion models such as Stable Diffusion \cite{rombach2022highresolution} and DALL-E 3 \cite{Dalle-3} has led to the growth of image synthesis frameworks that are able to generate high resolution photo-realistic images.
These models have been adopted widely in downstream applications such as video generation \cite{wu2023tuneavideo}, image editing \cite{hertz2022prompttoprompt}, robotics \cite{gao2023pretrained}, and more.
Multiple variations of T2I models have also been developed, which vary according to their text encoder \cite{chen2023pixartalpha}, priors \cite{ramesh2022hierarchical}, and inference efficiency \cite{luo2023latent}.
However, a common bottleneck that affects all of these methods is their inability to generate spatially consistent images: that is, given a natural language prompt that describes a spatial relationship, these models are unable to generate images that faithfully adhere to it. 

In this paper, we present a holistic approach towards investigating and mitigating this shortcoming through diverse lenses. We develop datasets, efficient training techniques, and explore multiple ablations and analyses to understand the behaviour of T2I models towards prompts that contain spatial relationships.

Our first finding reveals that existing vision-language (VL) datasets lack sufficient representation of spatial relationships.
Although frequently used in the English lexicon, we find that spatial words are scarcely found within image-text pairs of the existing datasets.
To alleviate this shortcoming, we create the ``SPRIGHT'' (\textbf{SP}atially \textbf{RIGHT}) dataset, the first spatially-focused large scale dataset.
Specifically, we synthetically re-caption $\sim$6 million images sourced from 4 widely used datasets, with a spatial focus (Section~\ref{sec:dataset}).
As shown in Figure \ref{fig:teaser}, SPRIGHT captions describe the fine-grained relational and spatial characteristics of an image, whereas human-written ground truth captions fail to do so.
Through a 3-fold comprehensive evaluation and analysis of the generated captions, we benchmark the quality of the generated captions and find that SPRIGHT largely improves over existing datasets in its ability to capture spatial relationships.
Next, leveraging only $\sim$0.25\% of our dataset, we achieve a 22\% improvement on the T2I-CompBench \cite{huang2023t2i} Spatial Score, and a 31.04\% and 29.72\% improvement in the FID \cite{heusel2017gans} and CMMD scores \cite{jayasumana2024rethinking}, respectively.

Our second finding reveals that significant performance improvements in spatial consistency of a T2I model can be achieved by fine-tuning on images that contain a large number of objects.
We achieve \textit{state-of-the-art} performance, and improve image fidelity, by fine-tuning on \textit{<500} image-caption pairs from SPRIGHT; training only on images that have a large number of objects. As investigated in VISOR \cite{gokhale2023benchmarking}, models often fail to generate the mentioned objects in a spatial prompt; we posit that by optimizing the model over images which have a large number of objects (and consequently, spatial relationships), we teach it to generate a large number of objects, which positively impacts its spatial consistency. In addition to improving spatial consistency, our model achieves large gains in performance across all aspects of T2I generation;  generating correct number of distinct objects, attribute binding and accurate generation in response to complex prompts.

We further demonstrate the impact of SPRIGHT by benchmarking the trade-offs achieved with long and short spatial captions, as well as spatially focused and general captions. We take the first steps towards discovering layer-wise activation patterns associated with spatial relationships, by examining the representation space of CLIP \cite{radford2021learning} as a text encoder. 

Our contributions and key findings are summarized below:

\begin{itemize}[nosep]
    \item
    We create SPRIGHT, the first spatially focused, large scale vision-language dataset by re-captioning $\sim$6 million images from 4 widely used existing datasets.To demonstrate the efficacy of SPRIGHT, we fine-tune baseline Stable Diffusion models on a small subset of our data and achieve performance gains across multiple spatial reasoning benchmarks while improving the corresponding FID and CMMD scores.
    \item
    We achieve state-of-the-art performance on spatial relationships by developing an efficient training methodology; specifically, we optimize over a small number (<500) of images which consists of a large number of objects, and achieve a 41\% improvement over our baseline model.
    \item
    Through multiple ablations and analyses, we present our findings related to spatial relationships: the impact of long captions, the trade-off between spatial and general captions, layer-wise activations of the CLIP text encoder, effect of training with negations and improvements over attention maps.
\end{itemize}
\section{Related Work}

\paragraph{\bf Text-to-image generative models.}
Since the initial release of Stable Diffusion \cite{rombach2022highresolution} and DALL-E \cite{ramesh2021zero}, different classes of T2I models have been developed, all optimized to generate highly realistic images corresponding to complex natural language prompts.
Models such as PixArt-Alpha \cite{chen2023pixartalpha}, Imagen \cite{saharia2022photorealistic}, and ParaDiffusion \cite{wu2023paragraphtoimage} move away from the CLIP text encoder, and explore traditional language models such as T5 \cite{raffel2020exploring} and LLaMA \cite{touvron2023llama} to process text prompts.
unCLIP \cite{ramesh2022hierarchical} based models have led to multiple methods \cite{patel2023eclipse, kakaobrain2022karlov1alpha} that leverage a CLIP-based prior as part of their diffusion pipeline.

\paragraph{\bf Spatial relationships in T2I models.}
Benchmarking the failures of T2I models on spatial relationships has been well explored by VISOR \cite{gokhale2023benchmarking}, T2I-CompBench \cite{huang2023t2i}, GenEval \cite{ghosh2023geneval}, and DALL-E Eval \cite{Cho2023DallEval}.
Both training-based and test-time adaptations have been developed to specifically improve upon these benchmarks.
Control-GPT \cite{zhang2023controllable} finetunes a ControlNet \cite{zhang2023adding} model by generating TikZ code representations with GPT-4 and optimizing over grounding tokens to generate images.
SpaText \cite{spa}, GLIGEN\cite{li2023gligen}, and ReCo \cite{yang2022reco} are training-based methods that introduce additional conditioning in their fine-tuning process to achieve better spatial control for image generation.
LLM-Grounded Diffusion \cite{lian2024llmgrounded} is a test-time multi-step method that improves over layout generated LLMs in an iterative manner.
Layout Guidance \cite{chen2023trainingfree} restricts objects to their annotated bounding box locations through refinement of attention maps during inference.
LayoutGPT \cite{feng2023layoutgpt} creates an LLM guided initial layout in the form of CSS, and then uses layout-to-image models to create indoor scenes.  REVISION \cite{chatterjee2024revisionrenderingtoolsenable} leverages 3D rendering engines to generate synthetic images which act as additional guidance during image synthesis for accurate depiction of spatial relationships in the generated image.

\paragraph{\bf Synthetic captions for T2I models.}
The efficacy of using descriptive and detailed captions has recently been explored by DALL-E 3 \cite{Dalle-3}, PixArt-Alpha \cite{chen2023pixartalpha} and RECAP \cite{segalis2023picture}.
DALL-E 3 builds an image captioning module by jointly optimizing over a CLIP and language modeling objective.
RECAP fine-tunes an image captioning model (PALI \cite{chen2023pali}) and reports the advantages of fine-tuning the Stable Diffusion family of models on long, synthetic captions.
PixArt-Alpha also re-captions images from the LAION \cite{laion} and Segment Anything \cite{kirillov2023segment} datasets; however their key focus is to develop descriptive image captions.
On the contrary, our goal is to develop captions that explicitly capture the spatial relationships seen in the image.
\section{The SPRIGHT Dataset} \label{sec:dataset}

We find that current vision-language (VL) datasets do not contain ``enough''  relational and spatial relationships.
Despite being frequently used in the English vocabulary \footnote{\url{https://www.oxfordlearnersdictionaries.com/us/wordlists/oxford3000-5000}}, words like ``left/right'', ``above/behind'' are scarce in existing VL datasets.
This holds for both annotator-provided captions, e.g., COCO \cite{lin2014microsoft}, and web-scraped alt-text captions, e.g., LAION \cite{laion}.
We posit that the absence of such phrases is one of the fundamental reasons for the lack of spatial consistency in current text-to-image models. Furthermore, language guidance is now being used to perform mid-level \cite{Yun_2023_CVPR, Xu_2023_CVPR} and low-level \cite{VPD, kondapaneni2024text} computer vision tasks. This motivates us to create the SPRIGHT (\textbf{SP}atially \textbf{RIGHT}) dataset, which explicitly encodes fine-grained relational and spatial information found in images.

\subsection{Creating the SPRIGHT Dataset}
We re-caption approximately six million images from four existing vision-language datasets, \ie datasets containing images and their corresponding natural language descriptions:
\begin{itemize}
    \item \textbf{CC-12M} \cite{cc12m} : 
    We re-caption a total of 2.3 million images from the CC-12M dataset, filtering out images of resolution less than 768 $\times$ 768.
    \item \textbf{Segment Anything (SA)} \cite{kirillov2023segment} : 
    We select Segment Anything as most images in it encapsulates a large number of objects; i.e. larger number of spatial relationships can be captured from a given image. We re-caption 3.5 million images as part of our re-captioning process.
    Since SA does not have ground-truth captions, we  generate its \textit{general} captions using the CoCa \cite{yu2022coca} model.
    
    \item \textbf{COCO} \cite{lin2014microsoft} : 
    We re-caption images ($\sim$ 40,000) from the validation set.
    
    \item \textbf{LAION-Aesthetics}\footnote{\url{https://laion.ai/blog/laion-aesthetics/}} : 
    We used 50,000 images from LAION-Aesthetics.
    \footnote{The entire LAION-5B dataset has been recalled for safety review: \url{https://laion.ai/notes/laion-maintenance/}. We will release our re-captioning outputs for these images based on the conclusions of this safety review.}
\end{itemize}
We use LLaVA-1.5-13B \cite{liu2023improved} with the following prompt to produce synthetic \textit{spatial captions} to create the SPRIGHT dataset: 
\begin{tcolorbox}[boxsep=0pt,boxrule=1pt]
    \scriptsize
    \texttt{Using 2 sentences, describe the spatial relationships seen in the image. You can use words like left/right, above/below, front/behind, far/near/adjacent, inside/outside. Also describe relative sizes of objects seen in the image.}
\end{tcolorbox}

\begin{table}[t]
    \caption{Compared to ground truth annotations, SPRIGHT consistently improves the presence of relational and spatial relationships captured in its captions, across diverse images from different datasets.}
    \centering
    \small
    \resizebox{\linewidth}{!}{
    \begin{tabular}{@{}l rcrcrcrcrcrcrcrcrcrcrc@{}}
        \toprule
        \multirow{2}{*}{\textbf{Dataset}} & \multicolumn{20}{c}{\textbf{\% of Spatial Phrases}} \\
         \cmidrule{2-23}
        && left && right && above && below &&  front &&  behind && next && close && far && small && large \\
            \midrule
        COCO &&  0.16 && 0.47 && 0.61 && 0.15 && 3.39 && 1.09 && 6.17 && 1.39 && 0.19 && 3.28 && 4.15 \\  
        \quad + SPRIGHT && 26.80 && 23.48 && 21.25 && 5.93 && 41.68 && 21.13 && 36.98 && 15.85 && 1.34 && 48.55 && 61.80 \\ 
        \midrule 
        CC-12M &&   0.61 && 1.45 && 0.40 && 0.19 && 1.40 && 0.43 && 0.54 && 0.94 && 1.07 && 1.44 && 1.44 \\
        \quad + SPRIGHT && 24.53 && 22.36 && 20.42 && 6.48 && 41.23 && 14.37 && 22.59 && 12.9 && 1.10 && 43.49 && 66.74 \\ 
        \midrule
        LAION && 0.27 && 0.75 && 0.16 && 0.05 && 0.83 && 0.11 && 0.24 && 0.67 && 0.91 && 1.03 && 1.01 \\
        \quad + SPRIGHT && 24.36 && 21.7 && 14.27 && 4.07 && 42.92 && 16.38 && 26.93 && 13.05 && 1.16 && 49.59 && 70.27 \\ 
        \midrule
         Segment Anything && 0.02 && 0.07 && 0.27 && 0.06 && 5.79 && 0.19 && 3.24 && 7.51 && 0.05 && 0.85 && 10.58 \\
         \quad + SPRIGHT && 18.48 && 15.09 && 23.75 && 6.56 && 43.5 && 13.58 && 33.02 && 11.9 && 1.25 && 52.19 && 80.22\\ 
         \bottomrule
    \end{tabular}
    }
    \label{tab:dataset_analysis_spatial}
\end{table}
\begin{figure}[t]
    \centering
    \includegraphics[width=0.9\linewidth]{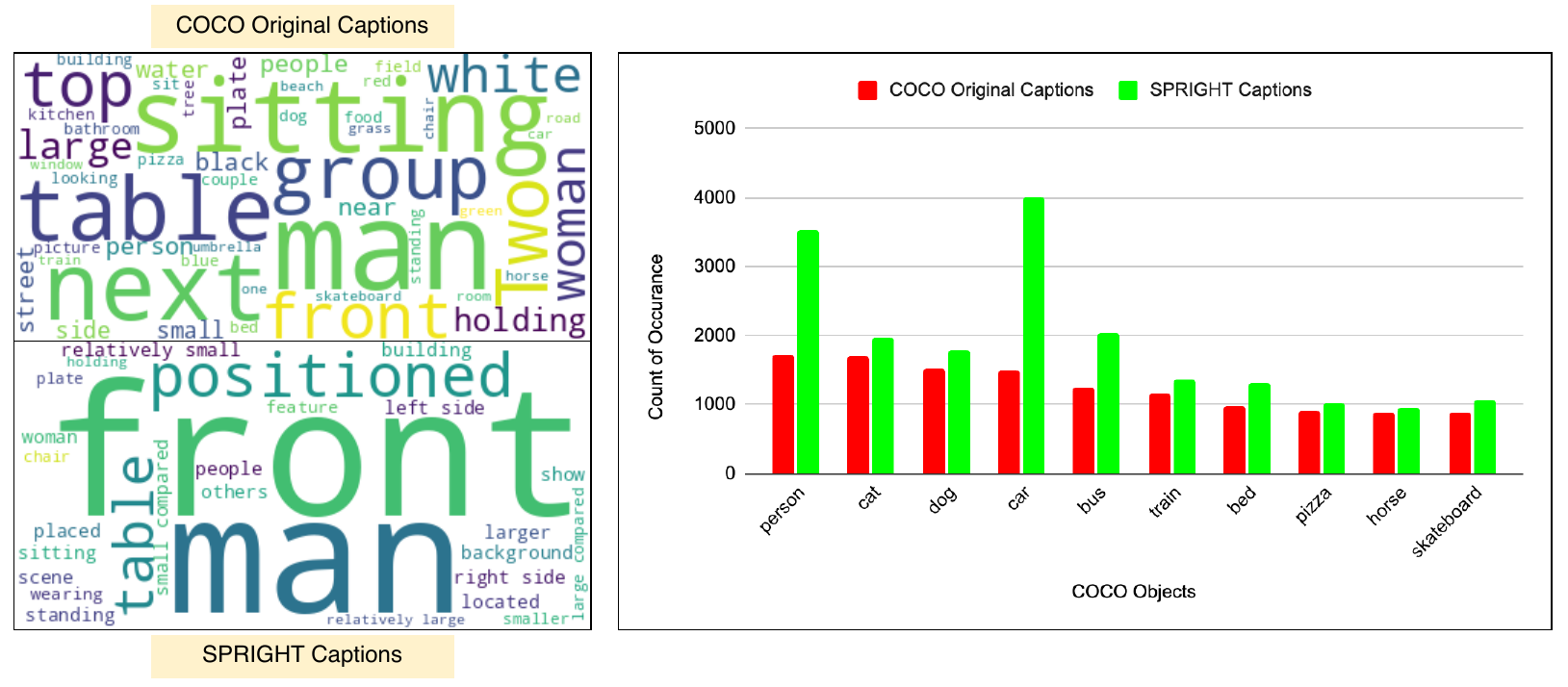}
    \caption{Compared to ground truth COCO captions,(\textbf{Left}) Word cloud representations showing that SPRIGHT captions significantly amplify the presence of spatial relationships. (\textbf{Right})SPRIGHT captions also capture a higher number of object occurances.}
    \label{fig:coco_image}
\end{figure}

\subsection{Impact of SPRIGHT}
Table \ref{tab:dataset_analysis_spatial} shows that SPRIGHT enhances the presence of spatial phrases across all relationship types on all the datasets.
\begin{table}
\caption{In addition to improving the presence of spatial relationships, SPRIGHT enhances linguistic diversity of captions in comparison to their original versions.}
\centering
\small
\resizebox{\linewidth}{!}{
\begin{tabular}{@{}lc cc cc cc cc@{}}
    \toprule
    \multirow{2}{*}{\textbf{Dataset}} && \multicolumn{7}{c}{\textbf{Average / caption}} \\
    \cmidrule{2-9}
    && Nouns && Adjectives && Verbs && Tokens \\
    \midrule
    COCO $\rightarrow$ COCO+SPRIGHT && 3.00 $\rightarrow$ 14.31 && 0.83 $\rightarrow$ 3.82 && 0.04 $\rightarrow$ 0.15 && 11.28 $\rightarrow$ 68.22 \\  
    \midrule
    CC-12M $\rightarrow$ CC-12M+SPRIGHT && 3.35 $\rightarrow$ 13.99 && 1.36 $\rightarrow$ 4.36 && 0.26 $\rightarrow$ 0.16 && 22.93 $\rightarrow$ 67.41 \\ 
    \midrule
    LAION $\rightarrow$ LAION+SPRIGHT && 1.78 $\rightarrow$ 14.32 && 0.70 $\rightarrow$ 4.53 && 0.11 $\rightarrow$ 0.14 && 12.49 $\rightarrow$ 69.74 \\ 
    \midrule
    SA $\rightarrow$ SA+SPRIGHT && 3.10 $\rightarrow$ 13.42 && 0.79 $\rightarrow$ 4.65 && 0.01 $\rightarrow$ 0.12 && 09.88 $\rightarrow$ 63.90 \\ 
    \bottomrule
\end{tabular}
}
\label{tab:dataset_analysis_general}
\end{table}
For 11 relationships, while the ground-truth captions of COCO and LAION only capture 21.05\% and 6.03\% of relationships, SPRIGHT captures 304.79\% and 284.7\%, respectively, \ie each re-captioned COCO image in SPRIGHT has $\sim$3 spatial phrases.
This shows that captions in VL datasets largely lack the presence of spatial relationships, and that SPRIGHT is able to improve upon this shortcoming by almost always capturing spatial relationships in every sentence.
Our captions offer several improvements beyond the spatial aspects:
\textbf{(i)} As depicted in Table \ref{tab:dataset_analysis_general} we improve the overall linguistic quality compared to the original captions, and \textbf{(ii)} we identify more objects and amplify their occurrences as illustrated in Figure \ref{fig:coco_image}; where we plot the top 10  objects present in the original COCO Captions and find that we significantly upsample their corresponding presence in SPRIGHT.

\subsection{Dataset Validation}
We perform 3 levels of evaluation to validate the SPRIGHT captions:\\

    \noindent\textbf{1. FAITHScore.}
    Following \cite{jing2023faithscore}, we leverage a large language model to deconstruct generated captions into atomic (simple) claims that can be individually and independently verified in a Visual Question Answering (VQA) format. We randomly sample 40,000 image-generated caption pairs from our dataset, and prompt GPT-3.5-Turbo to identify descriptive phrases (as opposed to subjective analysis that cannot be verified from the image) and decompose the descriptions into atomic statements. These atomic statements are then passed to LLaVA-1.5-13B for verification, and correctness is aggregated over 5 categories: entity, relation, colors, counting, and other attributes. We also measure correctness on spatial-related atomic statements, i.e., those containing one of the keywords left/right, above/below, near/far, large/small and background/foreground. The captions are on average 88.9\% correct, with spatially-focused relations, being 83.6\% correct; with the detailed breakdown presented in the Supplementary Materials. Since there is some uncertainty about bias induced by using LLaVA to evaluate LLaVA-generated captions, we also verify the caption quality in other ways, as described next.
    
    \smallskip\noindent\textbf{2. GPT-4 (V).} Inspired by recent methods \cite{Dalle-3, zhong2024multilora}, we perform a small-scale study on a split of 444 images from LAION and SA (from Section \ref{efficient_train}) to evaluate our captions with GPT-4(V) Turbo \cite{GPT-4(V)}. We prompt GPT-4(V) to rate each caption between a score of 1 to 10, especially focusing on the correctness of the spatial relationships captured. Captions of  images from LAION and SA had a \texttt{\{mean, median\}} rating of \texttt{\{7.49,8\}} and \texttt{\{7.36,8\}}, respectively. We present the prompt used in the Supplementary Materials.
    
    \smallskip\noindent\textbf{3. Human Annotation.}
    We also annotate a total of 3,000 images through a crowd-sourced human study, where each participant annotates a maximum of 30 image-text pairs. As evidenced by the average number of tokens in Table \ref{tab:dataset_analysis_spatial}, most captions in SPRIGHT have >1 sentences. Therefore, for fine-grained evaluation, we randomly select 1 sentence, from a caption in SPRIGHT, and evaluate its correctness for a given image. Across 149 responses, we find the metrics to be: \texttt{correct=1840} and \texttt{incorrect=928}, yielding an accuracy of 66.57\%.

\section{Improving Spatial Consistency} \label{improving}
In this section, we leverage SPRIGHT in an effective and efficient manner, and describe methodologies that significantly advance spatial reasoning in T2I models. 
We use Stable Diffusion v2.1 \footnote{\url{https://huggingface.co/stabilityai/stable-diffusion-2-1}} as the base model and our training and validation set consists of 13,500 and 1,500 images respectively, randomly sampled in a 50:50 split between LAION-Aesthetics and Segment Anything. 
Each image is paired with a typical caption and a spatial caption (from SPRIGHT). 
During fine-tuning, for each image, we randomly choose one of the given caption types in a 50:50 ratio. 
We fine-tune the U-Net and the CLIP text encoder as part of our training, both with a learning rate \(5 \times 10^{-6}\) optimized by AdamW \cite{loshchilov2018decoupled} and a global batch size of 128.
While we train the U-Net for 15,000 steps, the CLIP text encoder remains frozen during the first 10,000 steps. We develop our code-base on top of the Diffusers library \cite{von-platen-etal-2022-diffusers}.

\begin{table}[t]
    \caption{
    Quantitative metrics across multiple spatial reasoning and image fidelity metrics, demonstrating the effectiveness of high quality spatially-focused captions in SPRIGHT. 
    \textcolor{ForestGreen}{Green} indicates results of the model fine-tuned on SPRIGHT. For FID, we use cfg = 3.0 and 7.0 for the baseline and the fine-tuned model, respectively.}
    \centering
    \resizebox{\textwidth}{!}{
    \begin{tabular}{@{}l c c ccccc ccc@{}}
        \toprule
        \multirow{2}{*}{\textbf{Method}} & \multirow{2}{*}{\textbf{OA (\%)} ($\uparrow$)} & \multicolumn{6}{c}{\textbf{VISOR (\%) ($\uparrow$)}} & \multirow{2}{*}
        {\stackanchor{\textbf{T2I-CompBench ($\uparrow$)}}{\textbf{Spatial Score}}} &  \multirow{2}{*}{\textbf{ZS-FID ($\downarrow$)}} &  \multirow{2}{*}{\textbf{CMMD ($\downarrow$)}} \\
        
         \cmidrule{3-8}
        & & \textbf{uncond} & \textbf{{cond}} & \multicolumn{1}{c}{\textbf{1}} &  \multicolumn{1}{c}{\textbf{2}} &  \multicolumn{1}{c}{\textbf{3}} &  \multicolumn{1}{c}{\textbf{4}}\\
        \midrule
        SD 2.1  & 47.83 & 30.25 & 63.24 & 64.42 & 35.74 & 16.13 &  4.70 & 0.1507 & 21.646 & 0.703 \\
        \rowcolor{mygreen} ~ + SPRIGHT & 53.59 & 36.00  & 67.16 & 66.09 & 44.02 & 24.15 & 9.13 & 0.1840 & 14.925 & 0.494 \\
        \bottomrule
    \end{tabular}
    }
    \label{tab:visor_baseline_comp}
\end{table}

\begin{figure}[t]
    \centering 
    \includegraphics[width=\linewidth]{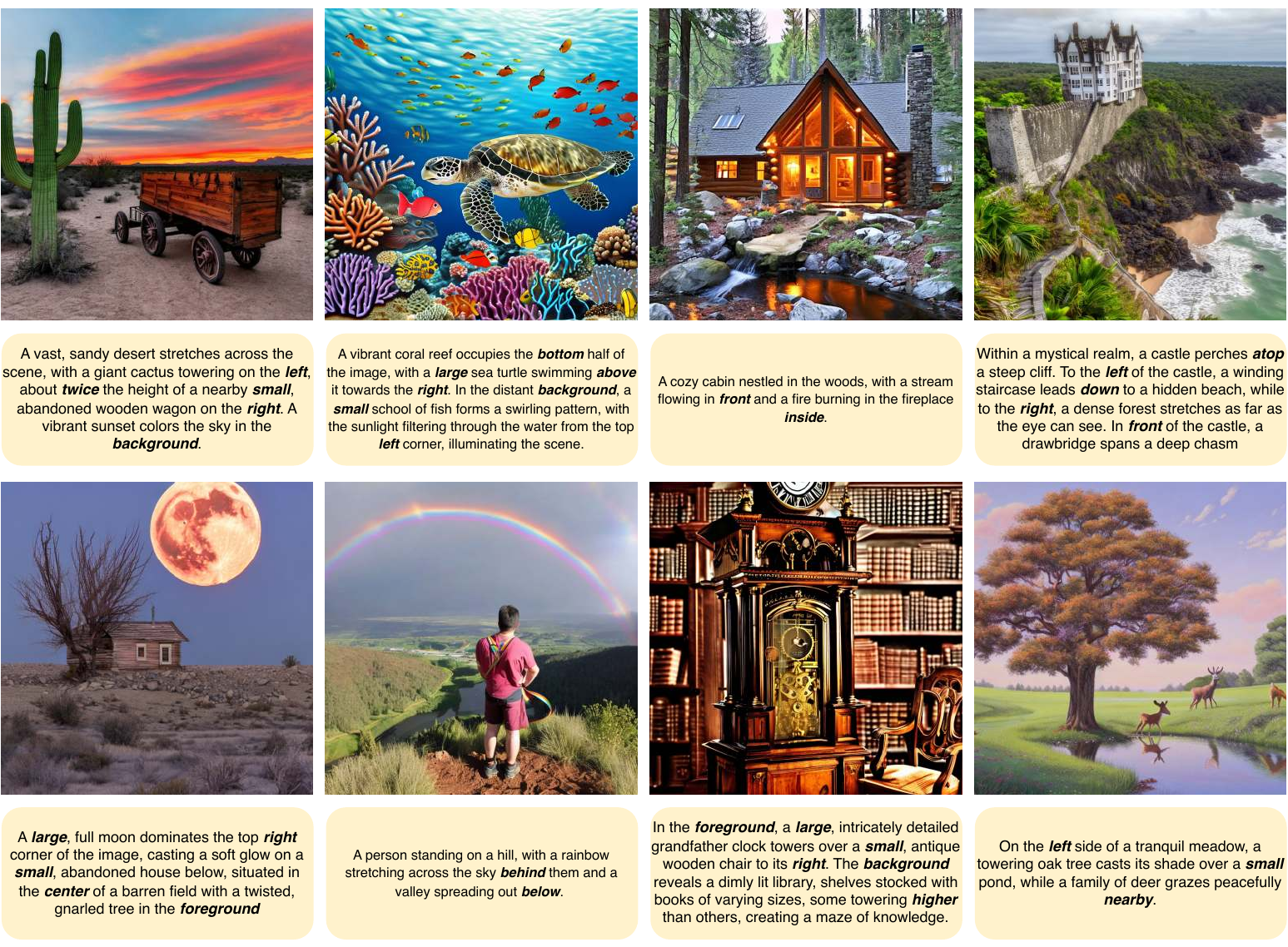}
    \caption{Generated images from our model, as described in Section \ref{baseline_improve}, on prompts which contain multiple objects and complex spatial relationships. We curate these prompts from ChatGPT.}
    \label{fig:complex_generated_images}
\end{figure}

\subsection{Improving upon Baseline Methods} \label{baseline_improve}

We present results on the spatial relationship benchmarks (VISOR \cite{gokhale2023benchmarking}, T2I-CompBench \cite{huang2023t2i}) and image fidelity metrics in Table \ref{tab:visor_baseline_comp}. 
To account for the inconsistencies associated with FID \cite{chong2020effectively, parmar2022aliased}, we also report results on CMMD \cite{jayasumana2024rethinking}. 
Across all metrics, our method significantly improves upon the base model by fine-tuning on <\textbf{\textit{15k}} images. We conclude that the dense, spatially focused captions in SPRIGHT provide effective spatial guidance to T2I models, and alleviate the need to scale up fine-tuning on a large number of images. As shown in Figure \ref{fig:complex_generated_images}, the model captures complex spatial relationships (\texttt{top right}), relative sizes (\texttt{large}) and patterns (\texttt{swirling}).

\begin{table}[t]
 \caption{Across all reported methods, we achieve \textit{state-of-the-art} performance on the T2I-CompBench Spatial Score. This is achieved by fine-tuning SD 2.1 on 444 image-caption pairs from the SPRIGHT dataset; where each image has >18 objects.}
 \centering

\begin{tabular}{@{}|l|rc rc rc rc a|}
    \hline
    \# of Objects per Image & {<6} && {<11} && {11} && {>11} && { > 18} \\
    \# of Training Images & 444 && 1346 && 1346 && 1346 && {444} \\
    T2I-CompBench Spatial Score ($\uparrow$) & 0.1309 && 0.1468 && 0.1667 && 0.1613 && {\textbf{0.2133}} \\
    \hline
\end{tabular}
\label{tab:num_obj_t2i_c}
\end{table}
\subsection{Efficient Training Methodology} \label{efficient_train}
We devise an additional efficient training methodology, which achieves state-of-the-art performance on the spatial aspect of the T2I-CompBench Benchmark.
We hypothesize that \textbf{\textit{(a)}} images that capture a large number of objects inherently also contain multiple spatial relationships; and \textbf{\textit{(b)}} training on these kinds of images will optimize the model to consistently generate a large number of objects, given a prompt containing spatial relationships; a well-documented failure mode of current T2I models \cite{gokhale2023benchmarking}. 

For our dataset of \textbf{{<15k}} images the median \# of objects/image = \textit{11}. 
We partition our dataset into multiple subsets based on the maximum number of objects present in an image.
\begin{table}[t]
    \caption{Comparing baseline SD 2.1 with our state-of-the-art model, across multiple spatial reasoning and image fidelity metrics, as described in Section \ref{efficient_train}.
    \textcolor{ForestGreen}{Green} indicates results from our model. 
    For FID, we use cfg = 3.0 and 7.5 for the baseline model and our model, respectively}
    \centering
    \resizebox{\linewidth}{!}{
    \begin{tabular}{@{}l c cccccc ccc@{}}
        \toprule
        \multirow{2}{*}{\textbf{Method}} & \multirow{2}{*}{\textbf{OA (\%)} ($\uparrow$)} & \multicolumn{6}{c}{\textbf{VISOR (\%) ($\uparrow$)}} & \multirow{2}{*}
        {\stackanchor{\textbf{T2I-CompBench ($\uparrow$)}}{\textbf{Spatial Score}}} &  \multirow{2}{*}{\textbf{ZS-FID ($\downarrow$)}} &  \multirow{2}{*}{\textbf{CMMD ($\downarrow$)}} \\
        
         \cmidrule{3-8}
        & & \textbf{uncond} & \textbf{{cond}} & \multicolumn{1}{c}{\textbf{1}} &  \multicolumn{1}{c}{\textbf{2}} &  \multicolumn{1}{c}{\textbf{3}} &  \multicolumn{1}{c}{\textbf{4}}\\
        \midrule
        SD 2.1  & 47.83 & 30.25 & 63.24 & 64.42 & 35.74 & 16.13 &  4.70 & 0.1507 & 21.646 & 0.703 \\
        \rowcolor{mygreen} + \stackanchor{SPRIGHT}{(<500 images)} & 60.68 & 43.23 & 71.24 & 71.78 & 51.88 & 33.09 & 16.15 & 0.2133 & 16.149 & 0.512  \\
        \bottomrule
    \end{tabular}
    }
    \label{tab:app_visor_baseline_comp}
\end{table}
This partitioning is automated using the open-world image tagging model Recognize Anything \cite{zhang2023recognize}.
We create five subsets, train corresponding models on a single subset and benchmark them in Table \ref{tab:num_obj_t2i_c}.
We keep the same hyper-parameters as before, only initiating training of the CLIP Text Encoder from the beginning.
With an increase in the \# of objects / image, iterative improvement in spatial fidelity is observed, with the best score for the subset containing greater than 18 objects.

Our major finding is that, with \textbf{444} training images and spatial captions from SPRIGHT, we achieve a 41\% improvement over the baseline SD 2.1 and attain state-of-the-art performance across all reported models on the T2I-CompBench spatial score. In Table \ref{tab:app_visor_baseline_comp}, compared to SD 2.1, we significantly improve all aspects of the VISOR score, while also enhancing the ZS-FID and CMMD scores on COCO-30K images by 25.39\% and 27.16\%, respectively. Our key findings on VISOR (Table \ref{tab:visor_main}) include: (a) a 26.86\% increase in the Object Accuracy (OA) score, indicating substantial gains in generating objects mentioned in the input prompt, and (b) a VISOR\(_{4}\) score of 16.15\%, demonstrating our model's consistent generation of spatially accurate images.

\begin{table}[t]
    \caption{\textbf{Results on the VISOR Benchmark}. Our model outperforms existing methods, on all aspects related to spatial relationships, consistently generating spatially accurate images as shown by the high VISOR [1-4] values.}
    \centering
    \small
    \begin{tabular}{@{}l rr r rrrr@{}}
        \toprule
        \multirow{2}{*}{\textbf{Method}} & \multirow{2}{*}{\textbf{OA (\%)}} & \multicolumn{6}{c}{\textbf{VISOR (\%)}} \\
         \cmidrule{3-8}
        & & \textbf{uncond} & \textbf{{cond}} & \multicolumn{1}{c}{\textbf{1}} &  \multicolumn{1}{c}{\textbf{2}} &  \multicolumn{1}{c}{\textbf{3}} &  \multicolumn{1}{c}{\textbf{4}}\\
        \midrule
        GLIDE   \cite{nichol2022glide}       &  3.36 &  1.98 & 59.06 &  6.72 & 1.02 & 0.17 & 0.03 \\ 
         GLIDE + CDM  \cite{liu2022compositional}       &  10.17 &  6.43 & 63.21 & 20.07 & 4.69 & 0.83 & 0.11 \\ 
         CogView2 \cite {ding2022cogview2}  & 18.47 & 12.17 & \underline{65.89} & 33.47 & 11.43&  3.22 & 0.57 \\ 
        DALLE-mini \cite{Dayma_DALL·E_Mini_2021}    & 27.10 & 16.17 & 59.67 & 38.31 & 17.50 &  6.89 & 1.96 \\
        DALLE-2  \cite{ramesh2022hierarchical}     & \textbf{63.93} & \underline{37.89}  & 59.27 & \textbf{73.59} & \underline{47.23} & \underline{23.26} & \underline{7.49} \\
        Structured Diffusion \cite{feng2023trainingfree} & 28.65 & 17.87 & 62.36 & 44.70 & 18.73 & 6.57 & 1.46 \\ 
        Attend-and-Excite \cite{chefer2023attend} & 42.07 & 25.75 & 61.21 & 49.29 & 19.33 & 4.56 & 0.08 \\ 
        \midrule
        Ours (<500 images) &  \underline{60.68} & \textbf{43.23} & \textbf{71.24} & \underline{71.78} & \textbf{51.88} & \textbf{33.09} & \textbf{16.15} \\ 
        \bottomrule
    \end{tabular} 
    \label{tab:visor_main}
\end{table}
\begin{table}[t]
    \caption{\textbf{Results on the GenEval Benchmark.} In addition to spatial relationships, we also improve model performance in generating the correct number of objects.}
    \centering
    \resizebox{\linewidth}{!}{
    \begin{tabular}{l cc cc cc cc cc cc cc}
        \toprule
        \textbf{Method} & \textbf{Overall} && \stackanchor{\textbf{Single}}{\textbf{object}} && \stackanchor{\textbf{Two}}{\textbf{objects}} && \textbf{Counting} && \textbf{Colors} && \textbf{Position} && \stackanchor{\textbf{Attribute}}{\textbf{binding}} \\
        \midrule
        CLIP retrieval \cite{beaumont-2022-clip-retrieval} & 0.35 && 0.89 && 0.22 && 0.37 && 0.62 && 0.03 && 0.00 \\
        minDALL-E \cite{min-dalle} & 0.23 && 0.73 && 0.11 && 0.12 && 0.37 && 0.02 && 0.01 \\
        SD 1.5 & 0.43 && 0.97 && 0.38 && 0.35 && 0.76 && 0.04 && 0.06 \\
        SD 2.1 & 0.50 && 0.98 && 0.51 && \underline{0.44} && \textbf{0.85} && 0.07 && \underline{0.17} \\
        SDXL \cite{podell2023sdxl} & \textbf{0.55} && \underline{0.98} && \textbf{0.74} && 0.39 && \textbf{0.85} && \textbf{0.15} && \textbf{0.23} \\
        PixArt-Alpha \cite{chen2023pixartalpha} & 0.48 && 0.98 && 0.50 && \underline{0.44} && \underline{0.80} && 0.08 && 0.07 \\
        \midrule
        Ours (<500 images) & \underline{0.51} && \textbf{0.99} && \underline{0.59} && \textbf{0.49} && \textbf{0.85} && \underline{0.11} && 0.15 \\
        \bottomrule
    \end{tabular}
    }
    \label{tab:app_geneval_results}
\end{table}
We also compare our model's performance on the GenEval \cite{ghosh2023geneval} benchmark (Table \ref{tab:app_geneval_results}), and find that in addition to improving spatial relationship (see \textit{Position}), our model shows improvement in generating 1 and 2 objects, along with the correct number of objects. Throughout our experiments, our training approach not only preserves but also enhances the \textit{non-spatial} aspects associated with a text-to-image model. Additional results and illustrations from VISOR and T2I-CompBench are provided in the Supplementary Materials.
\begin{table}[t]
    \caption{Comparing (a) the effect the percentage of spatial captions and (b) the effect of long and short spatial captions.}
    \resizebox{\linewidth}{!}{
        \centering
        \subfloat[T2I-CompBench Spatial Scores for models trained on varying ratios of spatial captions. Fine-tuning on a ratio of 50\% and 75\% of spatial captions yields optimal  results.]{
            \begin{tabular}{@{}cc@{}}
                \toprule
                \textbf{\% of spatial captions} &  \stackanchor{\textbf{T2I-CompBench}}{\textbf{Spatial Score}} ($\uparrow$) \\
                \midrule
                25 & 0.154 \\
                \rowcolor{mygreen} 50 & 0.178 \\
                75 & 0.161 \\
                 100 & 0.140 \\
                \bottomrule
            \end{tabular}
        }
        ~
        \subfloat[T2I-CompBench Spatial Scores for models trained on long and short \textit{spatial} captions. Across multiple setups, we find that longer spatial captions lead to better improvements in spatial consistency.]{
            \begin{tabular}{@{}lcc@{}}
                \toprule
                \multirow{2}{*}{\textbf{Model, Setup}} & \multicolumn{2}{c}{\stackanchor{\textbf{T2I-CompBench }}{\textbf{Spatial Score ($\uparrow$)  }}} \\
                 \cmidrule{2-3}
                & Long Captions & Short Captions \\
                    \midrule
                SD 1.5, w/o CLIP FT & \hl{0.0910} & 0.0708 \\
                SD 2.1, w/o CLIP FT & \hl{0.1605} & 0.1420 \\
                SD 2.1, w/ CLIP FT & \hl{0.1777} &  0.1230 \\
                \bottomrule
            \end{tabular}
        }
    }
    \label{tab:combined_ablation}
\end{table}

\section{Ablation Studies and Analyses}
To fully ascertain the impact of spatially-focused captions in SPRIGHT, we experiment with multiple nuances of our dataset and the corresponding T2I pipeline. 
Unless stated otherwise, the experimental setup identical to Section~\ref{improving}.

\begin{figure}[t]
    \centering 
    \includegraphics[width=\linewidth]{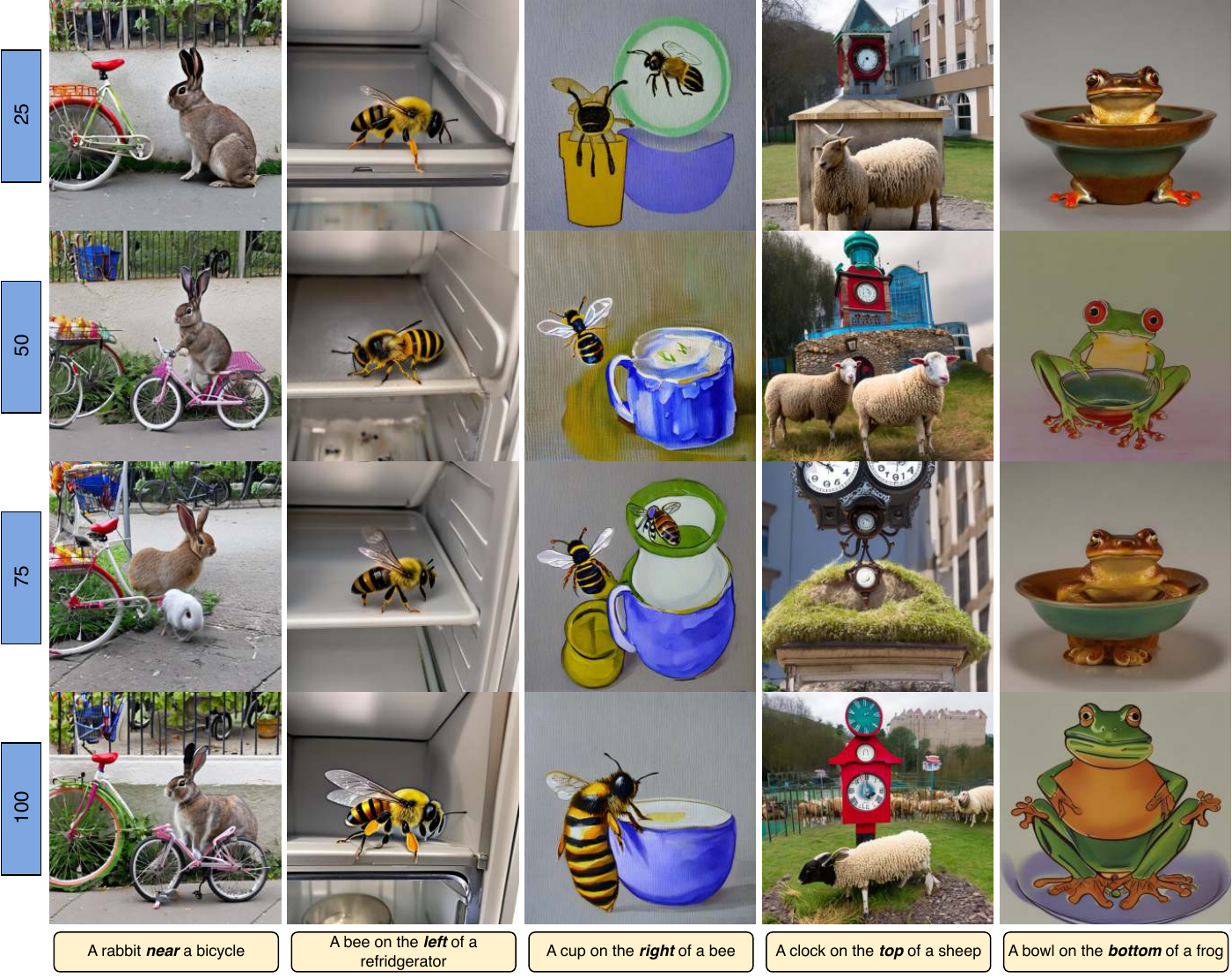}
    \caption{Illustrative comparisons between models trained on varying ratio of spatial experiments. Models trained on 50\% and 75\%  spatial captions are optimal.}
    \label{fig:ratio}
\end{figure}

\subsection{Optimal Ratio of Spatial Captions}
To understand the impact of spatially focused captions in comparison to ground-truth captions, we fine-tune different models by varying the \% of spatial captions. The results suggest that the model trained on 50\% spatial captions achieves the best spatial scores on T2I-CompBench (Table \ref{tab:combined_ablation} (a)). The models trained on only 25\% of spatial captions suffer largely from incorrect spatial relationships whereas the model trained only on spatial captions fails to generate the mentioned objects in the input prompt. Figure \ref{fig:ratio} shows illustrative examples.

\subsection{Impact of Long and Short Spatial Captions}
    
We also compare the effect of fine-tuning with shorter and longer variants of spatial captions. We create the shorter variants by randomly sampling 1 sentence from the longer caption, and fine-tune multiple models, with different setups. Across, all setups, (Table \ref{tab:combined_ablation} (b)) longer captions perform better than their shorter counterparts. In fact, CLIP fine-tuning hurts performance while using shorter captions, but has a positive impact on longer captions. 
This potentially happens because fine-tuning CLIP enables T2I models to generalize better to longer captions, which are out-of-distribution at the onset of training as they are initially pre-trained on short(er) captions from datasets such as LAION.

\begin{figure}[t]
    \centering 
    \includegraphics[width=\linewidth]{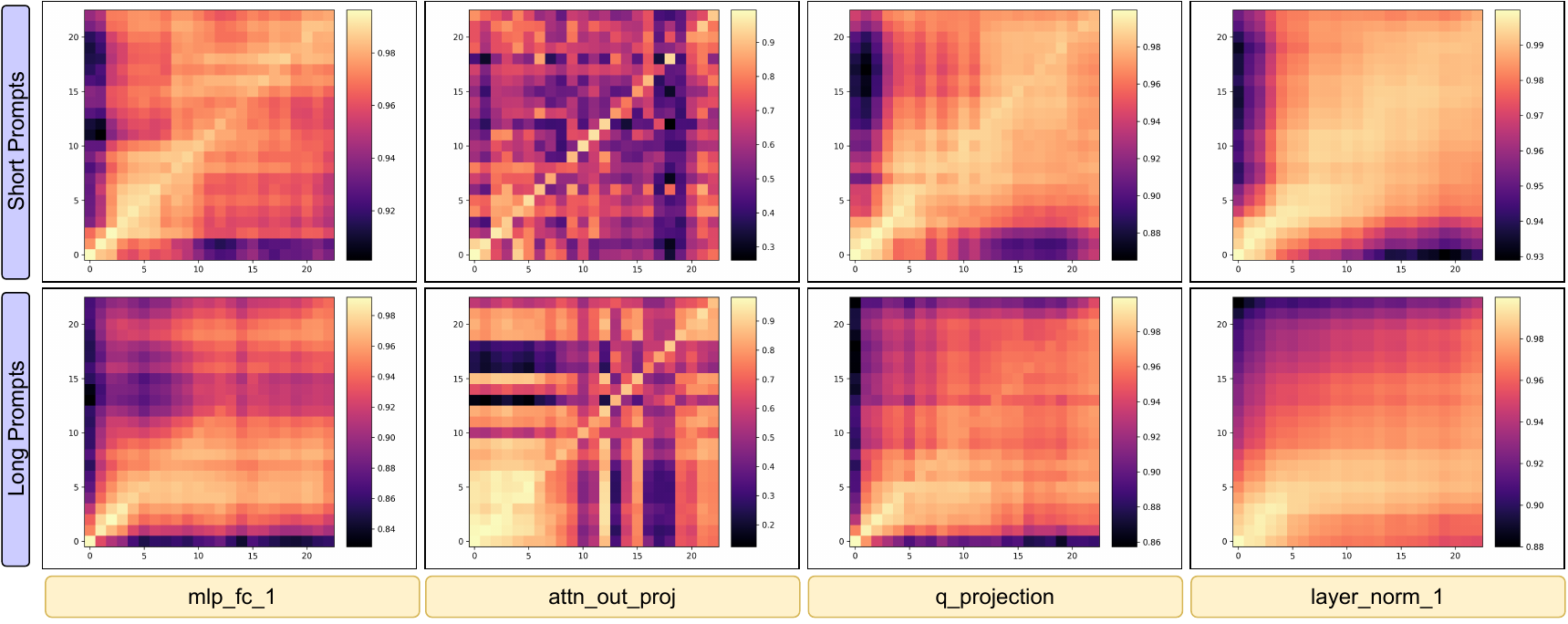}
    \caption{Comparison of layer-wise representations between Baseline CLIP (X-axis) and fine-tuned CLIP on SPRIGHT (Y-axis). Spatial captions show distinct representations in output attention projections and MLP layers, while layer norm layers are more similar. The representation gap widens with long, complex prompts, suggesting spatial prompts in SPRIGHT create diverse embeddings.}
    \label{fig:cka}
\end{figure}


\begin{table}[t]
    \caption{CLIP fine-tuned on SPRIGHT is able to differentiate the spatial nuances present in a textual prompt. While Baseline CLIP shows a high similarity for \textbf{\textit{spatially different}} prompts, SPRIGHT enables better fine-grained understanding.}
    \centering
    \small
    \resizebox{\linewidth}{!}{
    \begin{tabular}{lcccccc}
    \toprule
     & {``above''} & {``below''} & {``to the left of''} & {``to the right of''} & {``in front of''} & {``behind''} \\
    \midrule
    Baseline CLIP & 0.9225 & 0.9259 & 0.9229 & 0.9223 & 0.9231 & 0.9289 \\ 
    {CLIP + SPRIGHT} & 0.8674 & 0.8673 & 0.8658 & 0.8528 & 0.8417 & 0.8713 \\
    \bottomrule
    \end{tabular}
    }
    \label{tab:clip_exp}
\end{table}

\subsection{Investigating the CLIP Text Encoder}

The CLIP Text Encoder enables semantic understanding of the input text prompts in the Stable Diffusion model. As we fine-tune CLIP on the spatial captions, we investigate the various nuances associated with it: 

\paragraph{\bf Centered Kernel Alignment} (CKA) \cite{nguyen2021wide, kornblith2019similarity} compares layer-wise representations learned by two neural networks. Figure \ref{fig:cka} illustrates different representations learned by baseline CLIP, compared against the one trained on SPRIGHT. We compare layer activations across 50 simple and complex prompts  and aggregate representations from all the layers. Our findings reveal that the MLP and output attention projection layers play a larger role in enhancing spatial comprehension, as opposed to layers such as the layer norm. This distinction is larger with complex prompts, showing that the longer prompts from SPRIGHT indeed lead to more diverse embeddings being learned within the CLIP space.

\begin{figure}[t]
    \centering 
    \includegraphics[width=\linewidth]{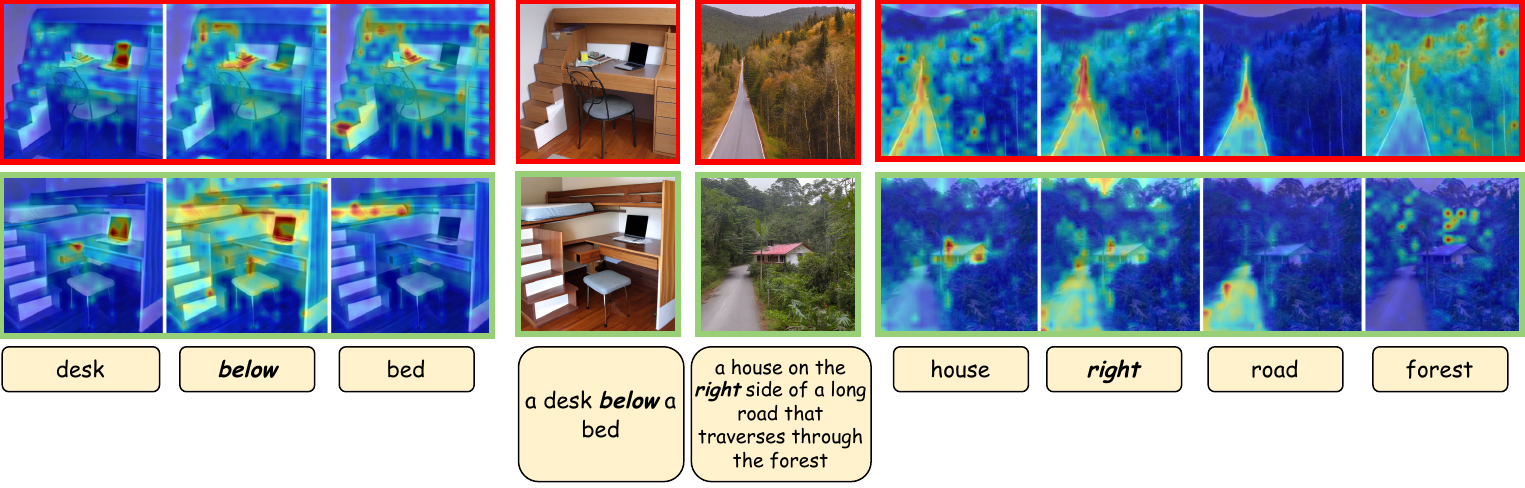}
    \caption{Visualising the cross-attention relevancy maps for baseline (\textbf{top row}) and fine-tuned model (\textbf{bottom row}) on SPRIGHT. Images in \textcolor{red}{red} are from baseline model while images in \textcolor{green}{green} are from our model.}
    \label{fig:attn_map}
\end{figure}

\paragraph{\bf Improving Semantic Understanding}: To evaluate semantic understanding of the fine-tuned CLIP, we perform the following experiment:  given a prompt containing a spatial phrase and 2 objects, we modify the prompt by switching the objects (\eg ``an airplane above an apple'' $\rightarrow$  ``an apple above an airplane''). 
Although these sentences have the same words, the placement of the two nouns relative to the preposition ``above'' completely changes the  meaning of the sentence.
To evaluate if models can discern this spatial distinction, we compute the cosine similarity between the pooled layer outputs of the original and modified prompts, for $\sim$ 37\textit{k} sentences. 
Table \ref{tab:clip_exp} shows that CLIP finetuned on SPRIGHT is able to differentiate between the prompts better (\ie lower cosine similarity) than the baseline.

\subsection{Improvement over Attention Maps}
Inspired by methods like Attend-and-Excite \cite{chefer2023attend}, we visualize attention relevancy maps for both simple and complex spatial prompts. Our model better generates the expected objects and achieves improved spatial localization compared to the baseline. For instance, the baseline models fails to generate objects like the bed and house, which our model successfully generates. The relevancy map indicates that high attention patches for missing words are spread across the image. Additionally, our model correctly attends to spatial words in the image, unlike the baseline. For example, in our model (Figure \ref{fig:attn_map}, bottom row), \texttt{below} attends to patches below the bed, and \texttt{right} attends to patches on the road's right, while Stable Diffusion 2.1 does not. We achieve these improvements across the intermediate attention maps and the final generated images.

\subsection{Training with Negation}
Dealing with negation remains a challenge for multimodal models as reported by previous findings on Visual Question Answering and Reasoning \cite{gokhale2020vqa, gokhale-etal-2022-semantically, dobreva-keller-2021-investigating}. Thus, in this section, we investigate the ability of T2I models to reason over spatial relationships and negations, simultaneously. Specifically, we study the impact of training a model with \texttt{``A man is not to the left of a dog''} as a substitute to \texttt{``A man is to the right of a dog''}. To create such captions, we post-process our generated captions and randomly replace spatial occurrences with their \textit{negation} counter-parts, and ensure that the semantic meaning of the sentence remains unchanged. Training on such a model, we find slight improvements in the spatial score, both while evaluating on prompts containing only negation (\textbf{\textit{0.069 > 0.066}}) and those that contain a mix of negation and simple statements (\textbf{\textit{0.1427 > 0.1376}}). There is however, a significant drop in performance, when evaluating on prompts that only contain negation; thus highlighting a major scope of improvement in this regard.
\section{Conclusion}

In this work, we present findings and techniques that enable improvement of spatial relationships in text-to-image models. We develop a large-scale dataset, SPRIGHT that captures fine-grained spatial relationships across a diverse set of images. Leveraging SPRIGHT, we develop efficient training techniques and achieve state-of-the art performance in generating spatially accurate images. 
We thoroughly explore various aspects concerning spatial relationships and evaluate the range of diversity introduced by the SPRIGHT dataset. We leave further scaling studies related to spatial consistency as future work. We believe our findings and results facilitate a comprehensive understanding of the interplay between spatial relationships and T2I models, and contribute to the future development of robust vision-language models.

\section*{Acknowledgements}
We thank Lucain Pouget for helping us in uploading the dataset to the Hugging Face Hub and the Hugging Face team for providing computing resources to host our demo. The authors acknowledge resources and support from the Research Computing facilities at Arizona State University. 
AC, CB, YY were supported by NSF Robust Intelligence program grants \#1750082 and \#2132724. TG was supported by Microsoft's Accelerating Foundation Model Research (AFMR) program and UMBC's Strategic Award for Research Transitions (START).
The views and opinions of the authors expressed herein do not necessarily state or reflect those of the funding agencies and employers.

\bibliographystyle{splncs04}
\bibliography{main}

\title{Getting it \textit{Right}: Improving Spatial Consistency in Text-to-Image Models : Supplementary Material} 

\titlerunning{Getting it \textit{Right}: Improving Spatial Consistency in Text-to-Image Models}

\author{
Agneet Chatterjee \thanks{Equal contribution. Correspondence to \href{mailto:agneet@asu.edu}{agneet@asu.edu}}\inst{1}\orcidlink{0000-0002-0961-9569} \and
Gabriela Ben Melech Stan \textsuperscript{$\star$} \inst{2}\orcidlink{0000-0001-6893-6647} \and
Estelle Aflalo\inst{2}\orcidlink{0009-0009-2860-6198} \and  \\
Sayak Paul\inst{3}\orcidlink{0000-0003-0217-0778} \and
Dhruba Ghosh \inst{4}\orcidlink{0000-0002-8518-2696} \and 
Tejas Gokhale \inst{5}\orcidlink{0000-0002-5593-2804} \and 
Ludwig Schmidt \inst{4} \and \\
Hannaneh Hajishirzi \inst{4}\orcidlink{0000-0002-1055-6657} \and 
Vasudev Lal\inst{2}\orcidlink{0000-0002-5907-9898} \and 
Chitta Baral\inst{1}\orcidlink{0000-0002-7549-723X} \and 
Yezhou Yang \inst{1}\orcidlink{0000-0003-0126-8976}
}

\authorrunning{A. Chatterjee et al.}

\institute{Arizona State University \and Intel Labs \and Hugging Face \and University of Washington \and University of Maryland, Baltimore County}

\maketitle

In this supplementary material, we present additional quantitave and qualitative results from our dataset and method. We discuss fine-grained FaithScore evaluations of the SPRIGHT captions, along with ways to improve the caption quality and its impact on models that support longer token limits. We present the GPT-4 (V) prompt used for evaluation and discuss the limitations of our current work. Lastly, we cover the contributions of each author in this work.

\definecolor{Gray}{gray}{0.85}
\newcolumntype{a}{>{\columncolor{Gray}}c}

\begin{table}[t]
\caption{\textbf{Results on the T2ICompBench Benchmark.} a) We achieve state of the art spatial score, across all methods, by efficient fine-tuning on only 444 images. b) Despite not explicitly optimizing for them, we find substantial improvement and competitive performance on attribute binding and non-spatial aspects.} 
\centering
\begin{tabular} {lcccac}
\toprule
\multicolumn{1}{c}
{\multirow{2}{*}{\bf Method}} & \multicolumn{3}{c}{\bf Attribute Binding } & \multicolumn{2}{c}{\bf Object Relationship} \\
\cmidrule(lr){2-4}\cmidrule(lr){5-6}
&
{\bf Color  } &
{\bf Shape} &
{\bf Texture} &
{\bf Spatial} &
{\bf Non-Spatial}
\\
\midrule
SD 1.4  & 0.3765 & 0.3576 & 0.4156 & 0.1246 & 0.3079   \\
SD 2 & 0.5065 & 0.4221 & 0.4922 & 0.1342 & 0.3096  \\
Composable v2 & 0.4063 & 0.3299 & 0.3645 & 0.0800 & 0.2980   \\
Structured v2  & 0.4990 & 0.4218 & 0.4900 & 0.1386 & 0.3111  \\
Attn-Exct v2  & 0.6400 & 0.4517 & 0.5963 & 0.1455 & 0.3109   \\
GORS  & {0.6603} & 0.4785 & 0.6287 & 0.1815 & {0.3193}  \\
DALLE-2  & 0.5750 & {0.5464} & {0.6374} & 0.1283 & 0.3043  \\
SDXL  & 0.6369 & 0.5408 & 0.5637 & {0.2032} & 0.3110  \\
PixArt-Alpha & {0.6886} & {0.5582} & {0.7044} & \underline{0.2082} & {0.3179}   \\
Kandisnky v2.2  & 0.5768 & 0.4999 & 0.5760 & 0.1912 & 0.3132 \\
DALL-E 3 & 0.8110 & 0.6750 & 0.8070 & - & - \\
\midrule
Ours (<500 images) & 0.6251 & 0.4648 & 0.5920 & \textbf{0.2133} & 0.3132 \\
\bottomrule
\end{tabular}

\label{tab:app_t2icompbench}
\end{table}
\begin{table}[!h]
\centering
\caption{FAITHScore caption evaluation of our SPRIGHT dataset. On a sample of 40,000 captions, SPRIGHT obtains an 88.9\% accuracy, comparable with the reported 86\% and 94\% on LLaVA-1k and MSCOCO-Captions, respectively. On the subset of atomic claims about spatial relations, SPRIGHT is correct 83.6\% of the time.}
\begin{tabular}{lcc}
    \toprule
    Category & \# Examples & Accuracy (\%) \\
    \midrule
    Overall & \multirow{2}{*}{---} & \multirow{2}{*}{88.9} \\
    \, FAITHScore & & \\
    \cmidrule{1-3}
    Entities & 149,393 & 91.4 \\
    Relations & 167,786 & 85.8 \\
    Colors & 10,386 & 83.1 \\
    Counting & 59,118 & 94.5 \\
    Other & 29,661 & 89.0 \\
    \cmidrule{1-3}
    Spatial & 45,663 & 83.6 \\
    \bottomrule
\end{tabular}
%

\label{tab:faithscore}
\end{table}
\section{Results on T2I-CompBench}
As shown in Table \ref{tab:app_t2icompbench}, we achieve state of the art performance on the spatial score in the widely accepted T2I-CompBench benchmark. The significance of training on images containing a large number of objects is emphasized by the enhanced performance of our models across various dimensions in T2I-CompBench. Specifically, we  enhance attribute binding parameters such as \textit{color} and \textit{texture}, alongside maintaining competitive performance in \textit{non-spatial} aspects.

\section{FaithScore Evaluations}

Table \ref{tab:faithscore} presents the detailed breakdown of the FaithScore evaluations conducted on the SPRIGHT captions, with the spatially-focused relationships being 83.6\% correct, on average.

\section{CLIP Token Limit}
The longer SPRIGHT captions better utilize the CLIP 77-token limit; ground truth and SPRIGHT captions have an average of \texttt{14.95} and \texttt{81.43} tokens, respectively. Furthermore, T2I models with longer context lengths and multiple text encoders such as PixArt-Sigma and SD3 can take full advantage of our captions and training technique: we fine-tune PixArt-Sigma (token limit = \texttt{300}) on SPRIGHT and obtain a spatial score of \texttt{0.2501}.

\section{Improvements in Captioning}
While our work is to explore the impact of spatially focused captions, we find that improvements in caption quality can  be achieved through stronger models like LLaVA-1.6-34B, GPT-4(V) or GPT-4o. To validate this, we conduct a human study ($n{=}3$) on 100 CC-12M images, comparing re-captioning performance of LLaVA-1.5-13B and LLaVA-1.6-34B, and find an improvement from 63\% to 78\%.

\section{System Prompt for GPT-4 Evaluation} 
\label{gpt_system_prompt}
\begin{tcolorbox}[]
    \footnotesize
    \texttt{You are part of a team of bots that evaluates images and their captions. Your job is to come up with a rating between 1 to 10 to evaluate the provided caption for the provided image. Consider the correctness of spatial relationships captured in the provided image. Return the response formatted as a dictionary with two keys: `rating', denoting the numeric rating, and `explanation', denoting a brief justification for the rating. \\\\
    The captions you are judging are designed to stress-test image captioning programs, and may include: \\
    1. Spatial phrases like above, below, left, right, front, behind, background, foreground (focus most on the correctness of these words). \\
    2. Relative sizes between objects such as small \& large, big \& tiny (focus on the correctness of these words). \\
    3. Scrambled or misspelled words (the image generator should produce an image associated with the probable meaning). Make a decision as to whether or not the caption is correct, given the image. \\\\
    A few rules: \\
    1. It is ok if the caption does not explicitly mention each object in the image; as long as the caption is correct in its entirety, it is fine. \\
    2. It is also ok if some captions don't have spatial relationships; judge them based on their correctness. A caption not containing spatial relationships should not be penalized. \\
    3. You will think out loud about your eventual conclusion. Don't include your reasoning in the final output. \\
    4. Return the response formatted as a Python-formatted dictionary having two keys: `rating', denoting the numeric rating, and `explanation', denoting a brief justification for the rating.}
\end{tcolorbox}

\section{Comparing COCO-30K and Generated Images}
\begin{figure}[t]
    \centering 
    \includegraphics[width=\linewidth]{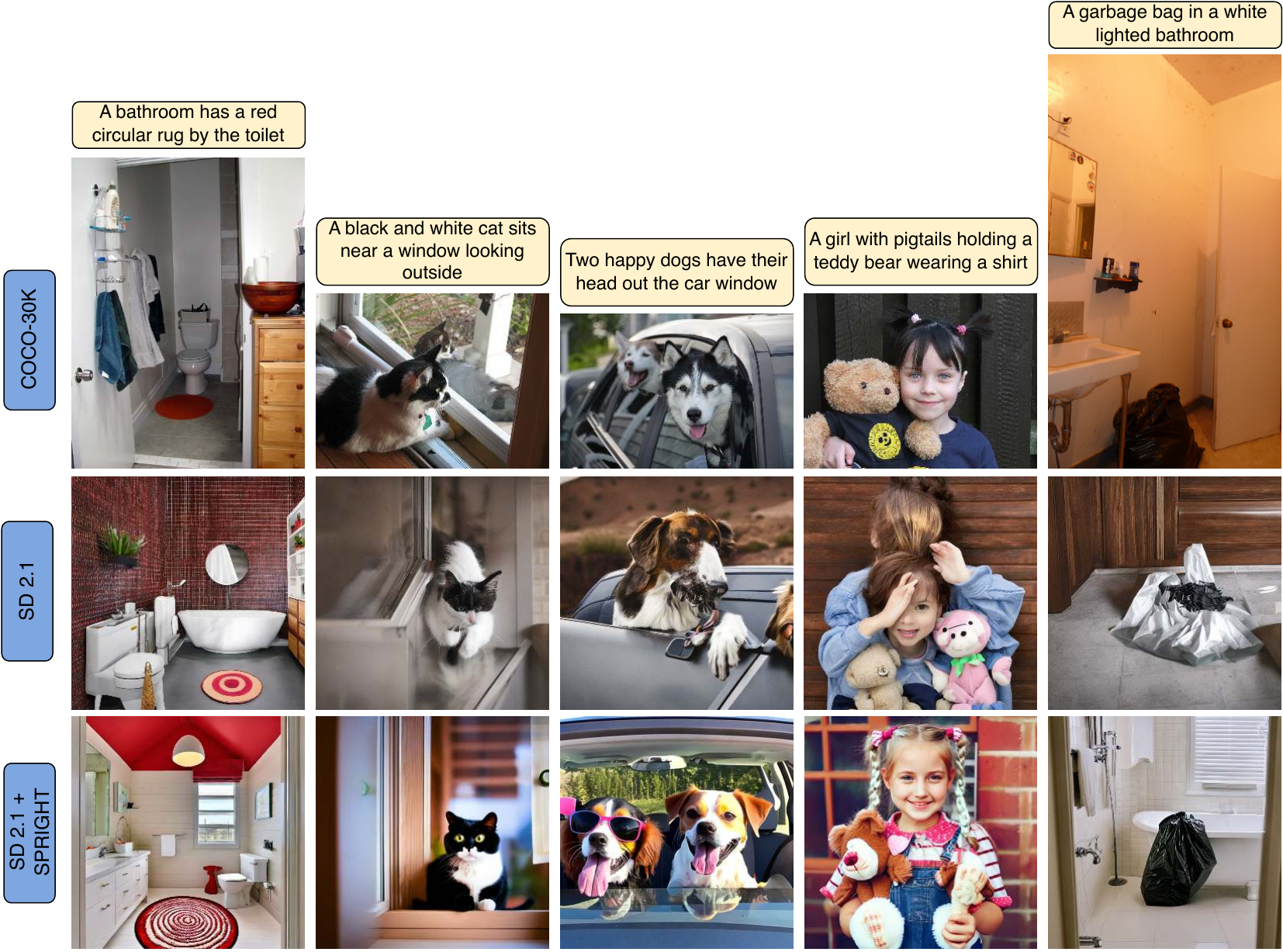}
    \caption{Illustrative examples comparing ground-truth images from COCO and generated images from Baseline SD 2.1 and our model. The images generated by our model exhibit greater fidelity to the input prompts, while also achieving a higher level of photorealism.}
    \label{fig:appendix_coco_30k}
\end{figure}

In Figure \ref{fig:appendix_coco_30k}, we compare images from COCO, baseline Stable Diffusion and our model. We find that the generated images from our model adhere to the input prompts better, are more photo-realistic in comparison to the baseline.

\section{Additional Examples from SPRIGHT}

\begin{figure}[t]
    \centering 
    \includegraphics[width=\linewidth]{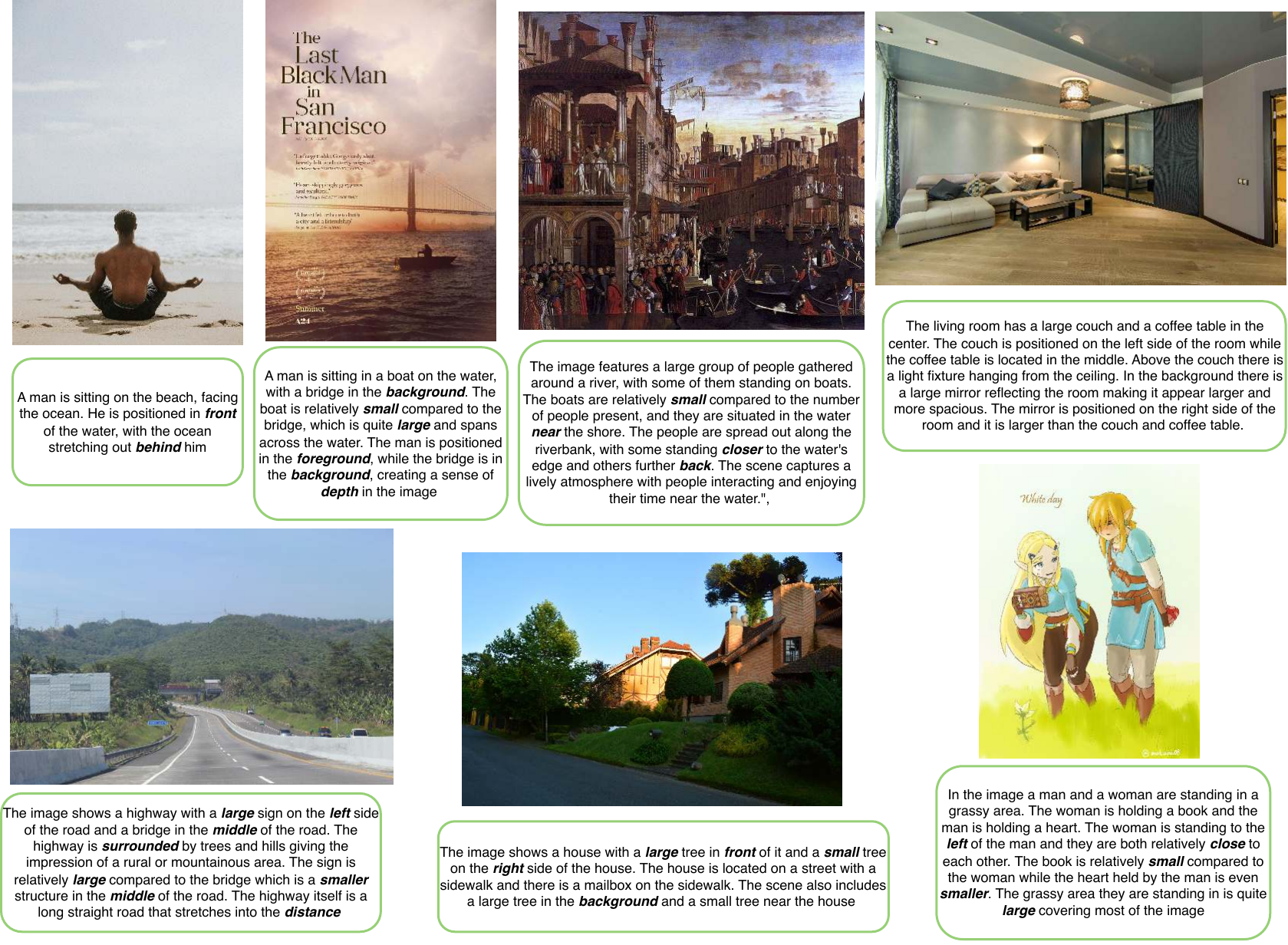}
    \caption{Illustrative examples from the SPRIGHT dataset, where the captions are correct in its entirety; both in capturing the spatial relationships and overall description of the image. The images are taken from CC-12M and Segment Anything. }
    \label{fig:spright_good}
\end{figure}

\begin{figure}[t]
    \centering 
    \includegraphics[width=\linewidth]{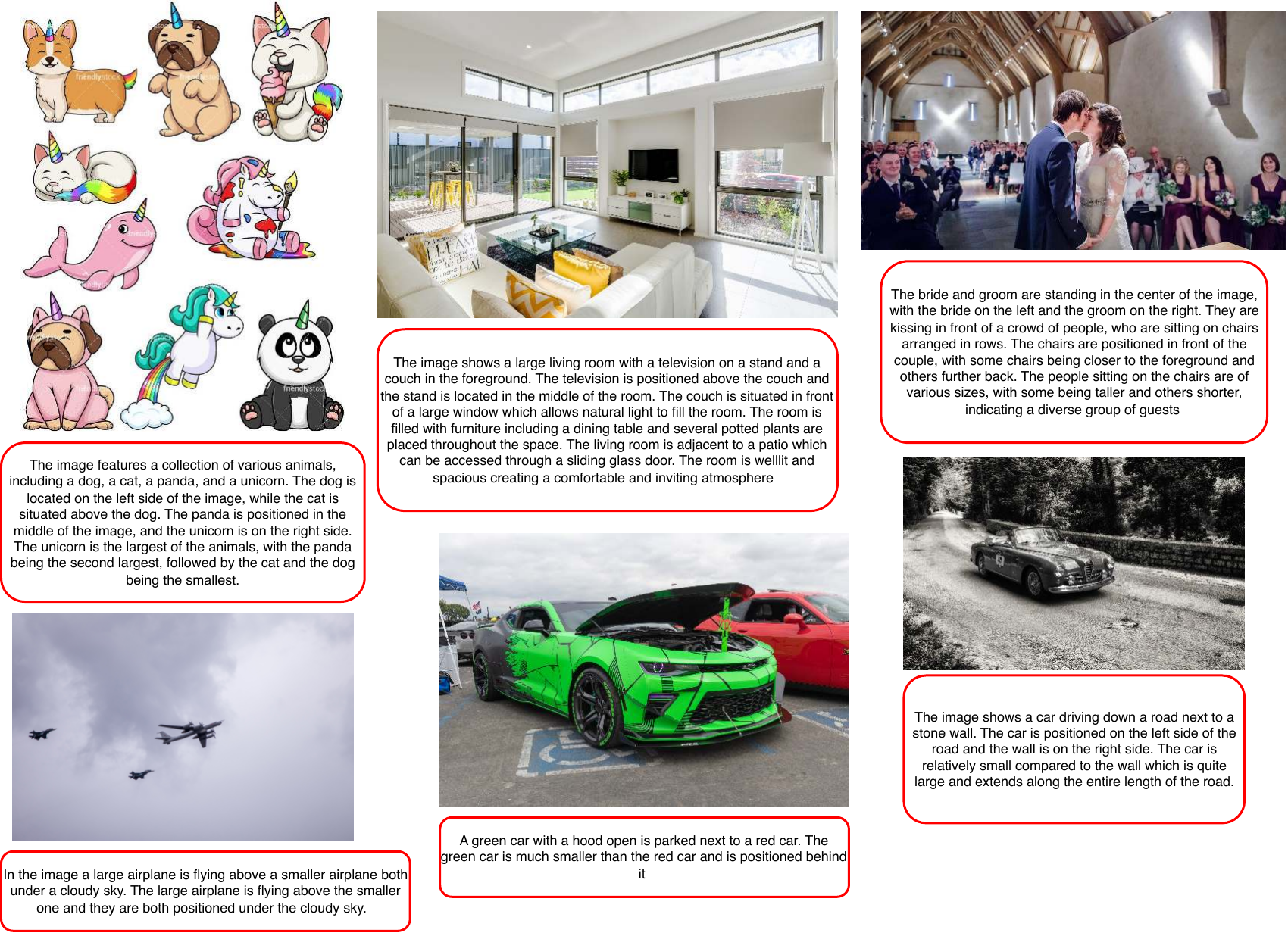}
    \caption{Illustrative examples from the SPRIGHT dataset, where the captions are not completely correct. The images are taken from CC-12M and Segment Anything.}
    \label{fig:spright_bad}
\end{figure}

Figure \ref{fig:spright_good} and \ref{fig:spright_bad} demonstrate a few correct and incorrect examples present in SPRIGHT. While most relationships are accurately described in the captions, on some instances the model struggles to capture the precise spatial nuance.

\begin{figure}[t]
    \centering 
    \includegraphics[width=\linewidth]{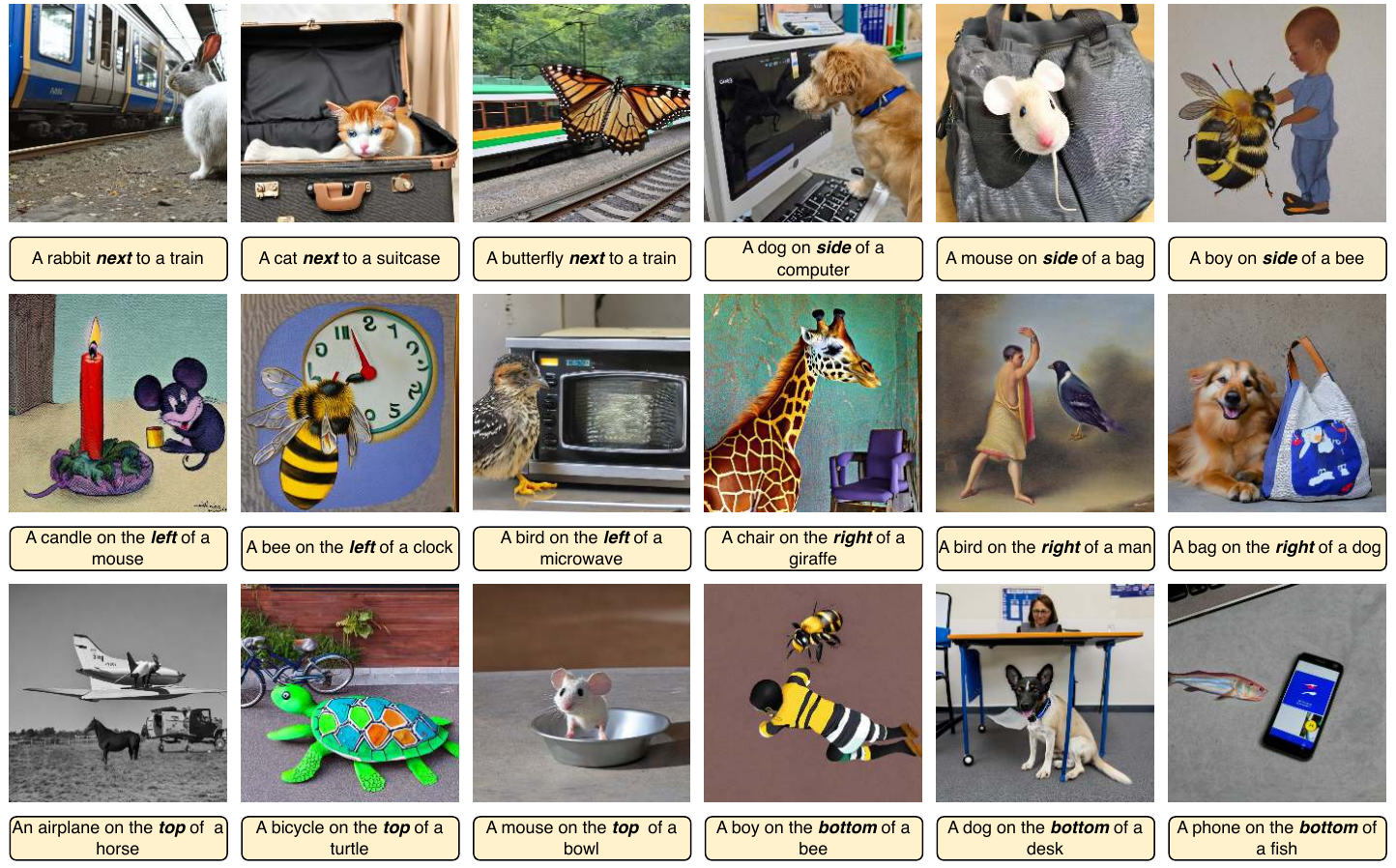}
    \caption{Illustrative examples from our model, as described in Section 4.1, on evaluation prompts from the T2I-CompBench benchmark.}
    \label{fig:t2i_c}
\end{figure}

\begin{figure}[t]
    \centering 
    \includegraphics[width=\linewidth]{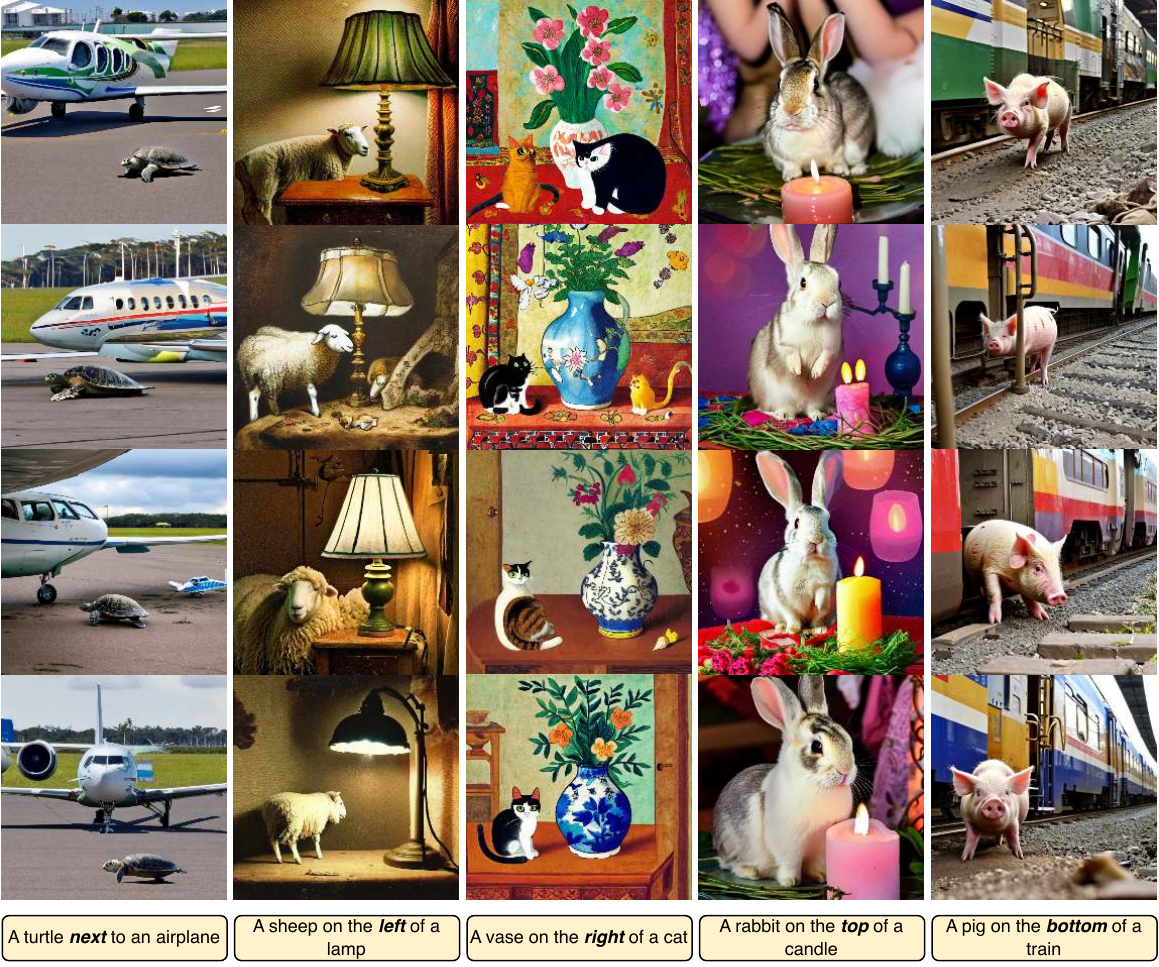}
    \caption{Generated images from our model, as described in Section 4.2, on evaluation prompts from T2I-CompBench. We find that for a given text prompt, our model consistently generates spatially accurate images.}
    \label{fig:consistency}
\end{figure}

\begin{figure}[t]
    \centering 
    \includegraphics[width=\linewidth,trim={5em 2em 5em 2em}]{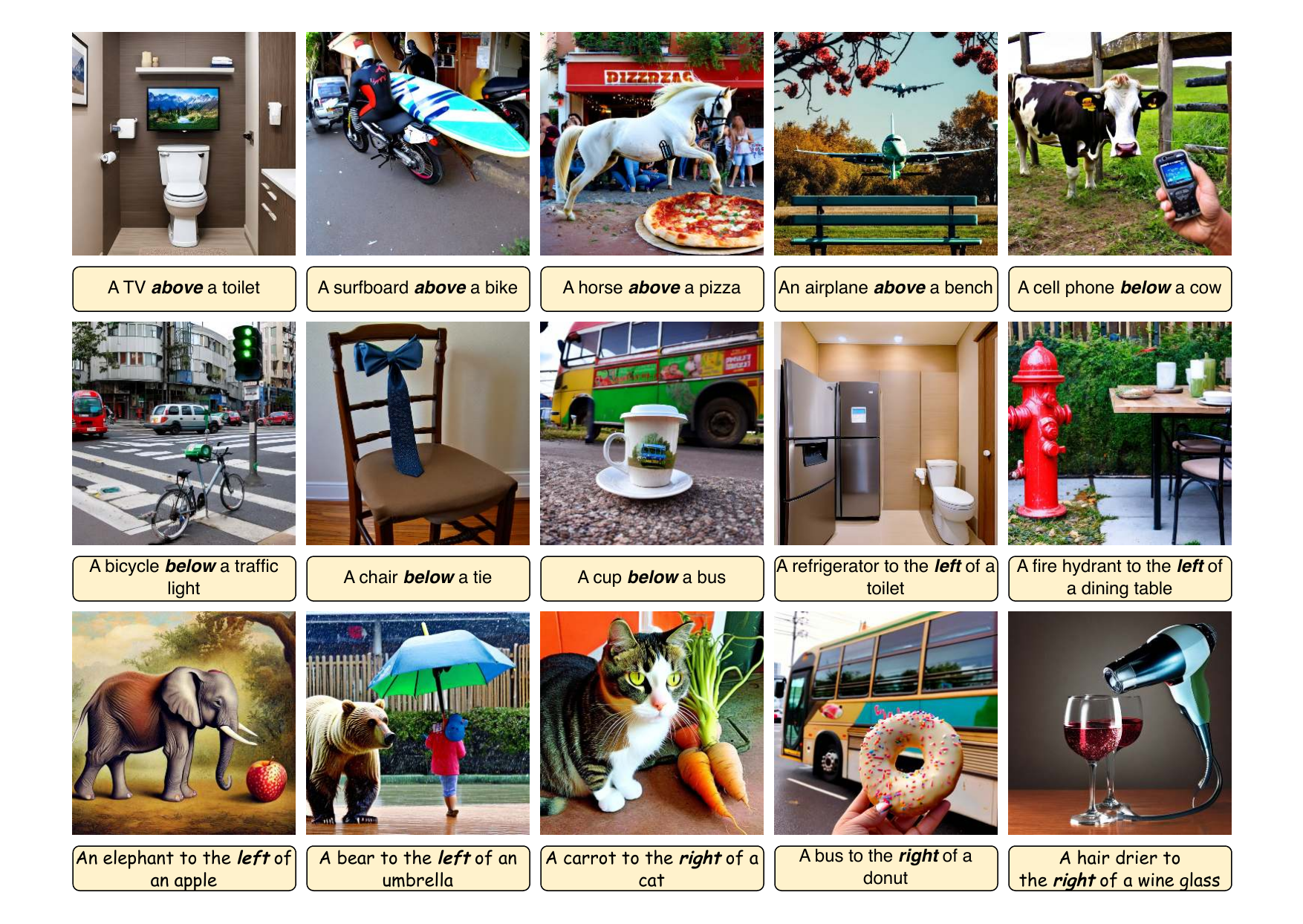}
    \caption{Generated images from our model, as described in Section 4.2, on evaluation prompts from the VISOR benchmark.}
    \label{fig:simple_visor_images}
\end{figure}

\section{Additional Illustrations}

Figure \ref{fig:t2i_c} shows images generated by our model based on prompts from T2I-CompBench, whereas Figure \ref{fig:consistency} demonstrates that for a given prompt, our model consistently produces spatially accurate images. Figure \ref{fig:simple_visor_images} presents example images generated from the VISOR benchmark.

\section{Limitations}
Since SPRIGHT is a derived dataset, it inherits the limitations of the original datasets. We refer the readers to the respective papers that introduced the original datasets for more details. As shown in our analysis, the generated synthetic captions are not a 100\% accurate and could be improved. The improvements can be achieved through better prompting techniques, larger models or by developing methods that better capture low-level image-text grounding. However, the purpose of our work is \textit{not} to develop the perfect dataset, it is to show the impact of creating such a dataset and its downstream impact in improving vision-language tasks. Since our models are a fine-tuned version of Stable Diffusion, they may also inherit their limitations in terms of biases, inability to generate text in images, errors in generating correct shadow patterns. We present our image fidelity metrics reporting FID on COCO-30K. COCO-30K is not the best dataset to compare against our images, since the average image resolutions in COCO is lesser than those generated by our model which are of dimension 768. Similarly, FID largely varies on image dimensions and has poor sample complexity; hence we also report numbers on the CMMD metric.

\section{Author Contributions}
AC defined the scope of the project, performed the initial hypothesis experiments and conducted the evaluations. GBMS led all the experimental work and customized the training code. EA generated the dataset, performed the dataset and relevancy map analyses. SP took part in the initial experiments, suggested the idea of re-captioning and performed few of the evaluations and analyses. DG suggested the idea of training with object thresholds and conducted the FAITHScore and GenEval evaluations. TG initiated the discussions on spatial failures of T2I models and provided consultation on experiments. VL, CB, and YZ co-advised the project, initiated and facilitated discussions, and helped shape the the goal of the project. AC and SP wrote the manuscript in consultation with TG, LW, HH, VL, CB, and YZ. All authors discussed the result and provided feedback for the manuscript.

\end{document}